\documentclass{bmvc2k}
\usepackage{multirow}
\usepackage{bm}

\title{On Fine-Tuned Deep Features for Unsupervised Domain Adaptation}

\addauthor{Qian Wang}{qian.wang173@hotmail.com}{1}
\addauthor{Toby P. Breckon}{toby.breckon@durham.ac.uk}{1,2}

\addinstitution{
 Department of Computer Science\\
 Durham University\\
 Durham, UK
}
\addinstitution{
 Department of Engineering\\
 Durham University\\
 Durham, UK
}

\runninghead{Wang, Breckon}{On Fine-Tuned Deep Features for UDA}

\begin{document}

\maketitle

\begin{abstract}
Prior feature transformation based approaches to Unsupervised Domain Adaptation (UDA) employ the deep features extracted by pre-trained deep models without fine-tuning them on the specific source or target domain data for a particular domain adaptation task. In contrast, end-to-end learning based approaches optimise the pre-trained backbones and the customised adaptation modules simultaneously to learn domain-invariant features for UDA. In this work, we explore the potential of combining fine-tuned features and feature transformation based UDA methods for improved domain adaptation performance. Specifically, we integrate the prevalent progressive pseudo-labelling techniques into the fine-tuning framework to extract fine-tuned features which are subsequently used in a state-of-the-art feature transformation based domain adaptation method SPL (Selective Pseudo-Labeling). Thorough experiments with multiple deep models including ResNet-50/101 and DeiT-small/base are conducted to demonstrate the combination of fine-tuned features and SPL can achieve state-of-the-art performance on several benchmark datasets.
\end{abstract}

\section{Introduction} \label{sec:intro}
Deep learning has been a dominant technique for many computer vision tasks (e.g., image classification, object detection, semantic segmentation, etc.) in many real-world applications. A deep model (e.g., deep Convolutional Neural Networks \cite{he2016deep} and Vision Transformers \cite{yang2021tvt,xu2021cdtrans}) usually has millions of parameters and training such a deep model requires sufficient training data (e.g., millions of annotated images). However, there exist situations where training data are limited or even unavailable. For example, in medical image processing, annotating data for training is non-trivial and cost-intensive. Training a deep model from scratch on a relatively small training data set cannot achieve satisfactory performance. A typical solution to this problem is transfer learning. A simple yet effective transfer learning technique is fine-tuning. Using a deep model pre-trained on a large dataset such as ImageNet and fine-tuning it on the training data of a particular task has been a de facto choice in many computer vision and image processing tasks.

In many real-world applications, however, there is no training data from the task domain (i.e. target domain) but abundant labelled data from a different relevant domain (i.e. source domain). To take advantage of the labelled data in the source domain, domain adaptation approaches have been proposed so that the task in the target domain can be better handled. In particular, Unsupervised Domain Adaptation (UDA) problems assume that: labelled source domain data and unlabelled target domain data are available during training;  the test data in the target domain are available for training although their ground truth labels are unavailable hence the problem is under the transductive learning setting; the source and target domains share the same set of classes (i.e. closed-set domain adaptation).

Existing UDA approaches can be roughly categorised into two groups \cite{wang2020data}: feature transformation based approaches and deep feature learning based approaches. The former type of approaches use features extracted by pre-trained deep models and learn linear projections to transform the deep features into a new feature space of good properties (e.g., discriminant, domain aligned). The latter type of approaches learn an end-to-end deep neural network consisting of modules for domain adaptation and the classifier. The former type of approaches have obvious limitations in that the deep models are only used as a feature extractor without being further fine-tuned on the data of specific tasks.

In this work, we aim at investigating the potential of feature transformation based approaches when the deep features are properly fine-tuned. Instead of designing complicated modules for domain adaptation, we employ a simple fine-tuning mechanism to update the pre-trained deep model using the data of target tasks. It is shown that fine-tuned deep features are indeed superior to the original deep features. The combination of fine-tuned deep features and feature transformation based UDA approaches can achieve comparable or better performance than state-of-the-art approaches.

\section{Related Work}\label{sec:related}
In this section, we give a brief review of prior works relating to ours. We first review recent approaches to UDA including both feature transformation based and deep learning based ones. Subsequently, we introduce the handling of batch normalisation layers in domain adaptation problems.

\subsection{Unsupervised Domain Adaptation} \label{sec:related_uda}
As mentioned before, we categorise UDA approaches into two groups: {feature transformation approaches} \cite{zhang2017joint, ghifary2017scatter, sun2017correlation, wang2018visual} and {deep feature learning approaches} \cite{ganin2015unsupervised, chen2018joint, pei2018multi, zhang2018collaborative}. 
Feature transformation approaches aim at learning mapping functions between source and target domains or from the source/target domain to a common subspace. The deep features are usually extracted from pre-trained models without fine-tuning. Such features are usually better

Deep feature learning approaches to domain adaptation take advantage of the powerful representation learning capability of CNN models. The objectives are usually to learn domain-invariant features from raw image data in source and target domains in an end-to-end learning framework. For example, the gradient reversal layer has been widely employed for domain adaptation \cite{ganin2015unsupervised,ganin2016domain, pei2018multi, zhang2018collaborative}. However, training such models with only labelled source data biases to the source domain and leads to marginally better or even worse performance than the baseline (e.g., the vanilla ResNet50 without specific domain adaptation modules) for target domain sample classification.  

Inspired by the success of transformers in vision tasks \cite{dosovitskiy2020image,touvron2021training}, recently new approaches to UDA based on transformer frameworks \cite{xu2021cdtrans,yang2021tvt} have been proposed. These transformer based UDA approaches have demonstrated significantly improved performance on benchmark datasets. In our work, we show that the features extracted by the transformer based models (whether fine tuned or not) outperform their CNN based counterparts for UDA. The fine tuned features can usually achieve even higher performance.   


\subsection{Batch Normalisation in Domain Adaptation} \label{sec:related_bn}
Batch normalisation layers \cite{ioffe2015batch} are widely used in many modern CNN models including ResNet \cite{he2016deep}, DenseNet \cite{huang2017densely} and EfficientNet \cite{tan2019efficientnet}. They need to be treated differently from other layers such as Convolutional layers and Dense/Fully-connected layers during fine-tuning \cite{peng2020fine,frankle2020training,kanavati2021partial}. Recall that the batch normalisation layer computes the output $\bm{y}$ of the input $\bm{x}$ in the following way:
\begin{equation}
    \label{eq:bn}
    \bm{y} = \gamma (\frac{\bm{x}-\mu}{\sigma+\epsilon})+\beta
\end{equation}
where $\gamma$ and $\beta$ are two trainable affine parameters; the mean $\mu$ and standard deviation $\sigma$ are estimated during training. In existing works, different strategies have been employed to tackle the batch normalisation layers for UDA problems. Kanavati et al. \cite{kanavati2021partial} demonstrate that fine-tuning only the batch norm affine parameters leads to similar performance as to fine-tuning all of the model parameters for domain adaptation. Li et al. \cite{li2018adaptive} replace the statistics of batch normalisation layers in the source domain with those in the target domain. As a result, the method is very simple and parameter-free in contrast to other UDA approaches. Romijnders et al. \cite{romijnders2019domain} use a domain agnostic normalisation layer which computes the statistics (i.e. $\mu$ and $\sigma$) based on the source domain and applies them to the target domain during inference. Klingner et al. \cite{klingner2022unsupervised} adapt the batch normalisation layer statistics by mixing the statistics from both domains. In our experiments, we fix the statistics from the pre-trained weights and directly apply them to both source and target domains during fine-tuning and inference.

\begin{figure}
    \centering
    \includegraphics[width=\textwidth]{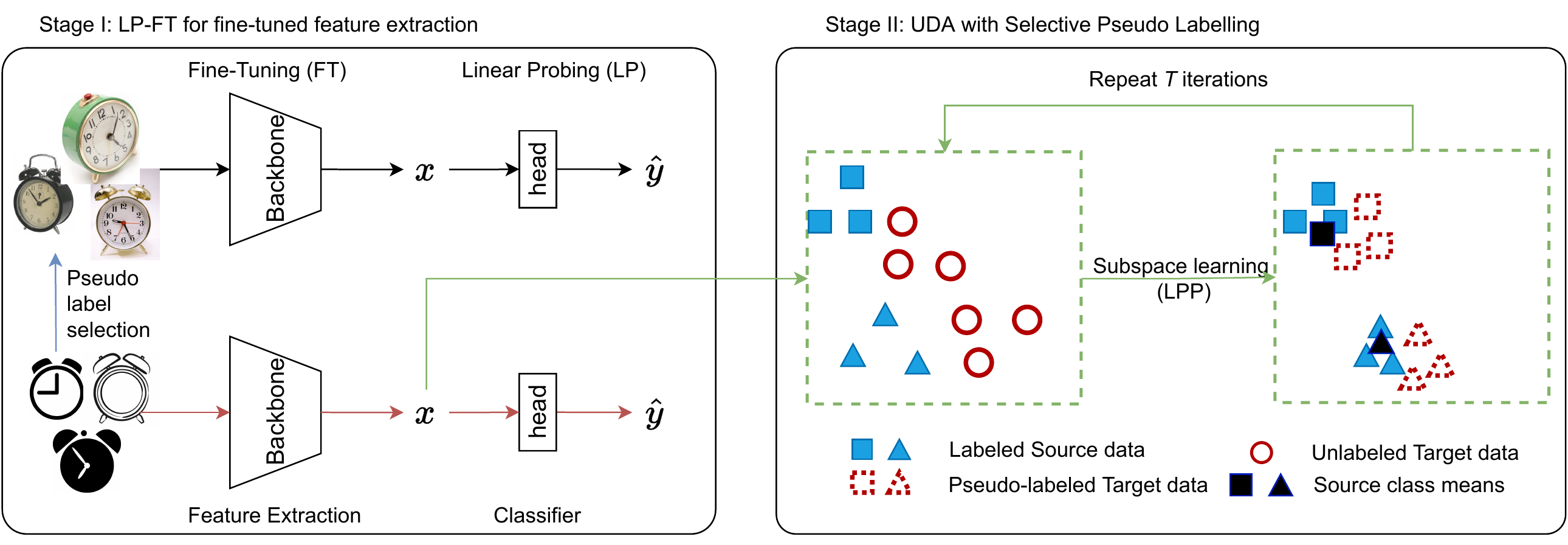}
    \caption{The framework of our proposed approach to UDA. The left box illustrates the fine-tuning process of the backbone model for feature extraction. The right side box illustrates how the fine-tuned features are used in the Selective Pseudo-Labeling (SPL) UDA method. }
    \label{fig:framework}
\end{figure}

\section{Method}\label{sec:method}
In this section, we present the details of how we fine-tune a pre-trained deep neural network and use the fine-tuned deep features to improve the unsupervised domain adaptation. Specifically, we first introduce the LP-FT (Linear Probing and Fine Tuning) scheme recently used in \cite{kanavati2021partial} and further analysed by Kumar et al. \cite{kumar2022fine}. Subsequently, we introduce how to take advantage of labelled source data and pseudo-labelled target data during fine-tuning. In our empirical study, two types of deep neural networks ResNet \cite{he2016deep} and DeiT \cite{touvron2021training} are considered. We describe a favourable trick needed to fine-tune networks containing batch normalisation layers (e.g., ResNet) for domain adaptation. Finally, to make the paper self-contained, we briefly describe the feature transformation based UDA approach SPL \cite{wang2020unsupervised} which is employed in our experiments.

\subsection{Linear Probing, Fine-Tuning and Batch Normalisation}\label{sec:lpft}
A deep model pre-trained on ImageNet has been effectively used to extract deep features for downstream tasks. Although a model pre-trained on a large-scale dataset like ImageNet can generate features of good generalisation, fine-tuning it on the training data for the specific task can usually further improve the performance. To fine-tune a model on the training data of a specific downstream task, one needs to replace the head of the model with a new one suitable for the task. The head of the model can be a linear layer containing the same number of neurons as the number of classes for a classification task. The model is initialised with the pre-trained weights from which the model weights are gradually updated based on the training data using an optimiser such as Stochastic Gradient Descent (SGD) and Adam.

During the fine-tuning, one can choose to update model weights of specific layers \cite{long2015learning,tajbakhsh2016convolutional,cetinic2018fine,guo2019spottune,vrbanvcivc2020transfer} rather than all layers of the network. Linear probing (LP) only updates the weights of the last classification layer and freezes all the rest weights for feature extraction \cite{chen2020generative,caron2021emerging,zhou2022ibot,kumar2022fine}. Recently, Kumar et al. \cite{kumar2022fine} demonstrate that LP-FT can outperform both linear probing and fine-tuning for out-of-distribution downstream tasks. LP-FT initialises the model with pre-trained weights and updates only the last classification layer (e.g. an L2 regularised logistic/softmax regression classifier) for some epochs. Subsequently, the whole network is fine tuned for weights in all layers to achieve the best performance \cite{kanavati2021partial,kumar2022fine}. 
It is also a popular strategy to fine-tune different layers with different learning rates \cite{peng2020fine,kanavati2021partial}. For ResNet models, we use a learning rate $\eta^{LP}=1e-2$ for the linear probing phase and adapt it to $\eta^{FT}=1e-4$ for all layers including the last linear classification layer in the fine-tuning phase. For DeiT models, we follow the same settings used in \cite{xu2021cdtrans}.

 In our employed LP-FT method, we set the batch norm layers trainable but in the inference mode. This means the trainable parameters in the batch norm layers (i.e. the coefficient $\gamma$ and the bias $\beta$ of the affine transformation) will be updated whilst the mean $\mu$ and standard deviation $\sigma$ will be frozen.

\subsection{Data for Fine-Tuning}\label{sec:data4ft}
For unsupervised domain adaptation, labelled source data are ready for use when fine-tuning the model. Fine-tuning the model with source data only enables the model to generate more discriminative features for source data but not necessarily for target data. Inspired by prior works on unsupervised domain adaptation \cite{wang2020unsupervised}, we select pseudo-labelled target samples progressively and add them to the training data set during fine-tuning. Specifically, after each epoch of fine-tuning, we select the most confidently predicted pseudo labels from the target data set. The confidence is based on the posterior probability $p(\hat{y}|x^t; \theta)$ output by the softmax layer. We employ classwise selection so that the numbers of selected pseudo-labelled target samples are balanced across all classes. In the following epoch of fine-tuning, the selected target samples along with their pseudo labels will be combined with the labelled source samples to form the new training data set. The number of selected pseudo-labelled target samples is linearly increased until all target samples are selected in the last epoch. In the early epochs of training, a small fraction of pseudo-labelled target samples are selected to participate in the fine-tuning to avoid the incorrect pseudo labels being reinforced in the following epochs. In the last epoch of fine-tuning, all target samples are exploited although incorrect pseudo labels still exist (unless the accuracy is 100\%).

\subsection{Method for UDA}\label{sec:method4uda}
To evaluate the fine-tuned features, we choose one feature transformation based UDA method in our study. To make the paper self-contained, we briefly describe the UDA approach Selective Pseudo Labelling (SPL) proposed in \cite{wang2020unsupervised} and more details can be found in the original paper.  The crucial components of SPL are subspace learning and selective pseudo labelling. A supervised Locality Preserving Projection (LPP) algorithm is employed to learn a common subspace into which both source and target data are projected. The learned subspace has favourable properties that projected samples in it are expected to be class discriminative and domain invariant. The subspace is learned based on the labelled source data and selected pseudo-labelled target samples. The pseudo labelled target samples are selected in a similar way to that used during model fine-tuning, i.e., the classwise selection is based on the pseudo label confidence and the number of selected pseudo labelled samples is linearly increased till the last iteration when all target samples are used for the subspace learning.

As for the classification, SPL uses the combination of Nearest Class Mean (NCM) \cite{wang2020unsupervised} and the Structured Prediction (SP) algorithms. NCM calculates the class means in the learned subspace and assigns the nearest class to the target samples. SP takes advantage of the cluster structure of target samples in the learned subspace and finds the optimal one-to-one match between the clusters and classes.

\section{Experiments and Results} \label{sec:expRes}
We conduct experiments on two benchmark datasets Office31 and Office-Home to show how the fine-tuned deep features can improve the UDA performance. We consider CNN models (i.e. ResNet50 and ResNet101 \cite{he2016deep}) and transformer based models (i.e. DeiT-small and DeiT-base \cite{touvron2021training}) in our experiments. We will describe the details of datasets, experimental settings, experimental results and compare the results with other state-of-the-art methods in the following subsections.

\subsection{Datasets}\label{sec:dataset}
\textbf{Office31} \cite{saenko2010adapting} consists of 4,110 images of 31 classes in three domains: Amazon (A), Webcam (W) and DSLR (D). Six domain adaptation tasks are used for the evaluation.
\textbf{Office-Home} \cite{venkateswara2017deep} consists of four domains: Artistic images (A), Clipart (C), Product images (P) and Real-World images (R). There are a total number of 15,588 images and 65 common object classes in four domains.

\subsection{Experimental Settings}\label{sec:setting}

We conduct experiments with the following methods.
\begin{itemize}
    \item \textbf{noFT+SPL}: the model pre-trained on ImageNet is used as a feature extractor to generate deep features (i.e. the activations of the second last layer) fed into the UDA approach SPL.
    \item \textbf{SourceOnlyFT}: the LP-FT scheme (c.f. Section \ref{sec:lpft}) is applied to the model and only the source data are used; the customized head (i.e. the last linear layer) serves as the domain-invariant classifier for the recognition of target samples and outputs the final results. 
    \item \textbf{SourceOnlyFT+SPL}: the fine-tuned model (i.e. SourceOnlyFT) is used as a feature extractor to generate deep features (i.e. the activations of the second last layer) fed into the UDA approach SPL. 
    \item \textbf{SourceTargetFT}: the LP-FT scheme (c.f. Section \ref{sec:lpft}) is applied to the model and the source data plus the progressively pseudo labelled target data are used; more specifically, the source data are used for the first 10 epochs of linear probing and the pseudo labelled target samples are progressively added for the fine-tuning from epoch 11 to 20; the customized head (i.e. the last linear layer) serves as the domain-invariant classifier for the recognition of target samples and outputs the final results.
    \item \textbf{SourceTargetFT+SPL}: the fine-tuned model (i.e. SourceTargetFT) is used as a feature extractor to generate deep features (i.e. the activations of the second last layer) fed into the UDA approach SPL.
\end{itemize}

\subsection{Experimental Results}\label{sec:results}
The experimental results of the dataset Office31 are shown in Table \ref{table:uda_o31}. The classification accuracies of six adaptation tasks, as well as the average accuracy of six tasks, are reported. A few conclusions can be drawn from the results in Table \ref{table:uda_o31}. First, when we use the customized head (i.e. the last linear layer for classification) as the classifier to output the final recognition results, the methods ``SourceOnlyFT" and ``SourceTargetFT" perform significantly worse than their counterparts with SPL. Second, when using SPL as the UDA approach, fine tuned deep features perform better than those extracted directly from the pre-trained model without fine-tuning. Whilst this is true for models like ResNet50 and ResNet101, there is no strong evidence for the models DeiT-small and DeiT-base. Third, the method ``SourceTargetFT" outperforms ``SourceOnlyFT", however, when the extracted features by these fine tuned models show no difference when fed into the UDA approach SPL, i.e., the method ``SourceTargetFT+SPL" is no better than ``SourceOnlyFT+SPL".

As for the experimental results on the Office-Home dataset (Table \ref{table:uda_o65}), the first two conclusions we draw above still hold. However, the third conclusion needs to be revised since ``SourceTargetFT" outperforms ``SourceOnlyFT" in most cases whether the SPL approach is used or not. 

The experimental results shown in Tables \ref{table:uda_o31} and \ref{table:uda_o65} provide evidence that properly fine tuned features are usually beneficial to UDA performance, especially when they are used in the feature transformation based UDA approaches such as SPL. In addition, using both source and target data in fine-tuning is usually superior to using source data only.

\begin{table}[!t]
	\centering
	{
		\centering
		\caption[]{Classification Accuracy (\%) on Office31 dataset using varying models and methods.
		}
		\label{table:uda_o31}
		\resizebox{.8\columnwidth}{!}{%
			\begin{tabular}{clccccccc}
				\hline
				Model & Method & \scriptsize{A$\to$W} & \scriptsize{D$\to$W} & \scriptsize{W$\to$D} & \scriptsize{A$\to$D} & \scriptsize{D$\to$A} & \scriptsize{W$\to$A} & Avg \\ \hline
                \multirow{5}{*}{ResNet50} & noFT+SPL & 92.7 & {98.7} & {99.8} & {93.0} & {76.4} & {76.8} & {89.6} \\
                & SourceOnlyFT & 76.4 & 96.2 & 99.8 & 79.3 & 69.6 & 67.7 & 81.5 \\
				& SourceOnlyFT+SPL & 93.0 & 98.6 & 99.8 & 95.0 & 78.0 & 77.2 & \bf 90.3 \\
				& SourceTargetFT & 81.9 & 97.0 & 99.8 & 82.9 & 68.7 & 66.7 & 82.8\\
				& SourceTargetFT+SPL & 92.3 & 98.7 & 99.8 & 93.8 & 78.5 & 77.4 & 90.1 \\
				\hline
				\multirow{5}{*}{ResNet101} & noFT+SPL & 90.1 & 99.0 & 99.8 & 92.4 & 79.8 & 78.5 & 89.8 \\
				&SourceOnlyFT & 80.9 & 97.0 & 99.4 & 83.1 & 70.8 & 69.0 & 83.4\\
				&SourceOnlyFT+SPL & 93.2 & 99.0 & 99.8 & 94.4 & 80.3 & 78.3 & \bf 90.8 \\
				&SourceTargetFT & 83.1 & 93.0 & 98.2 & 87.8 & 69.5 & 68.3 & 83.3\\
				&SourceTargetFT+SPL & 93.1 & 99.0 & 99.8 & 94.8 & 79.5 & 78.4 & \bf 90.8\\
				\hline
				\multirow{5}{*}{DeiT-small} & noFT+SPL & 95.5 & 98.6 & 100.0 & 96.2 & 78.5 & 80.5 & \bf 91.6\\
				& SourceOnly & 87.2 & 98.1 & 100.0 & 87.1 & 75.6 & 73.8 & 87.0\\
				& SourceOnly+SPL & 94.0 & 98.4 & 100.0 & 96.4 & 79.3 & 79.3 & 91.2 \\
				& SourceTargetFT & 93.6 & 98.2 & 100.0 & 95.0 & 75.6 & 75.3 & 89.6 \\
				& SourceTargetFT+SPL & 95.6 & 98.1 & 100.0 & 96.6 & 78.0 & 79.5 & 91.3\\
				\hline
				\multirow{5}{*}{DeiT-base} & noFT+SPL & 96.9 & 99.1 & 100.0 & 96.4 & 82.4 & 81.0 & 92.6\\
				& SourceOnlyFT & 89.4 & 98.5 & 100.0 & 91.0 & 76.2 & 75.4 & 88.4\\
				& SourceOnlyFT+SPL & 97.9 & 99.1 & 100.0 & 97.6 & 82.6 & 81.1 & 93.0 \\
				& SourceTargetFT & 94.7 & 98.4 & 100.0 & 96.6 & 77.1 & 76.5 & 90.6\\
				& SourceTargetFT+SPL & 97.6 & 99.1 & 100.0 & 98.0 & 81.3 & 82.4 & \bf 93.1\\
				\hline
			\end{tabular}%
		}
	}
\end{table}

\begin{table*}[!htbp]
	\centering
	{
		\centering
		\caption[]{Classification Accuracy (\%) on Office-Home dataset using varying models and methods.}
		\label{table:uda_o65}
		\resizebox{\columnwidth}{!}{%
			\begin{tabular}{clccccccccccccc}
				\hline
				Model & Method & A$\to$C & A$\to$P & A$\to$R & C$\to$A&C$\to$P & C$\to$R&P$\to$A & P$\to$C & P$\to$R & R$\to$A & R$\to$C & R$\to$P & Average \\ \hline
				\multirow{5}{*}{ResNet50} & noFT+SPL  & 53.9 & 78.6 & 82.1 & 65.2 & 78.2 & 81.0 & 66.3 & 52.5 & 82.9 & 70.5 & 55.4 & 85.9 & 71.0 \\
				& SourceOnlyFT & 44.3 & 67.2 & 74.4 & 55.1 & 66.9 & 67.5 & 55.0 & 42.2 & 74.3 & 64.2 & 44.9 & 76.4 &  61.0\\
				& SourceOnlyFT+SPL & 55.6 & 79.9 & 82.0 & 66.2 & 78.4 & 80.1 & 66.4 & 52.0 & 82.3 & 71.8 & 57.4 & 85.8 & 71.5 \\
				& SourceTargetFT & 43.0 & 66.7 & 73.9 & 58.2 & 67.8 & 68.8 & 55.8 & 43.4 & 76.3 & 64.5 & 43.5 & 76.6 & 61.5\\
				& SourceTargetFT+SPL & 55.9 & 79.3 & 82.3 & 67.1 & 79.6 & 79.4 & 67.0 & 54.8 & 82.1 & 71.6 & 56.7 & 85.7 & \bf 71.8 \\ 
				\hline
				\multirow{5}{*}{ResNet101} & noFT+SPL  & 57.5 & 81.0 & 84.0 & 68.3 & 79.2 & 81.9 & 68.0 & 56.4 & 83.9 & 74.1 & 59.5 & 87.3 & 73.4 \\
				& SourceOnlyFT & 48.3 & 69.2 & 76.0 & 57.8 & 67.7 & 70.5 & 55.6 & 44.6 & 75.6 & 66.6 & 48.1 & 78.8 &  63.2\\
				& SourceOnlyFT+SPL & 59.5 & 81.5 & 84.2 & 70.7 & 80.4 & 81.5 & 67.0 & 56.7 & 83.6 & 74.2 & 59.1 & 87.1 & \bf 73.8 \\
				& SourceTargetFT & 50.6 & 70.8 & 76.3 & 61.6 & 68.8 & 71.6 & 56.7 & 46.6 & 76.9 & 67.1 & 46.8 & 79.0 &  64.4 \\
				& SourceTargetFT+SPL & 59.2 & 82.5 & 83.8 & 71.3 & 80.4 & 81.8 & 68.1 & 55.4 & 83.7 & 73.4 & 58.3 & 87.7 & \bf 73.8\\
				\hline
				\multirow{5}{*}{DeiT-small} & noFT+SPL  & 58.5 & 85.6 & 85.2 & 73.3 & 84.2 & 84.6 & 71.0 & 58.5 & 85.5 & 76.3 & 59.3 & 87.2 & 75.8\\
				& SourceOnly & 56.4 & 75.7 & 81.7 & 71.0 & 75.0 & 78.0 & 67.7 & 52.0 & 81.8 & 74.3 & 53.2 & 83.7 & 70.9 \\
				& SourceOnly+SPL & 62.7 & 82.9 & 85.1 & 76.5 & 84.1 & 83.8 & 75.0 & 55.1 & 85.4 & 76.8 & 58.4 & 87.0 & 76.1 \\
				& SourceTargetFT & 59.6 & 77.3 & 82.9 & 73.6 & 78.4 & 80.3 & 71.3 & 56.7 & 83.7 & 73.4 & 54.8 & 85.3 & 73.1 \\
				& SourceTargetFT+SPL & 63.8 & 82.9 & 85.2 & 76.2 & 84.4 & 83.8 & 75.2 & 61.5 & 85.8 & 77.4 & 60.0 & 87.2 & \bf 77.0 \\
				\hline
				\multirow{5}{*}{DeiT-base} & noFT+SPL  & 63.4 & 86.5 & 87.5 & 78.6 & 86.1 & 86.2 & 72.8 & 59.3 & 87.8 & 77.6 & 61.9 & 89.5 & 78.1 \\
				&SourceOnlyFT & 60.9 & 79.2 & 84.1 & 73.4 & 78.6 & 80.7 & 71.0 & 55.1 & 84.3 & 77.6 & 58.4 & 85.8 & 74.1\\
				&SourceOnlyFT+SPL & 67.8 & 86.8 & 87.8 & 81.0 & 85.3 & 86.8 & 74.8 & 59.6 & 88.7 & 80.6 & 63.9 & 90.2 & 79.4 \\
				&SourceTargetFT & 62.0 & 80.6 & 85.6 & 75.7 & 82.0 & 82.7 & 73.4 & 56.6 & 85.8 & 77.0 & 60.4 & 87.0 & 75.7\\
				&SourceTargetFT+SPL & 69.4 & 85.1 & 87.9 & 81.2 & 85.7 & 86.5 & 78.0 & 64.4 & 89.0 & 81.0 & 66.7 & 90.6 & \bf 80.5 \\
				\hline
			\end{tabular}%
		} 
	}
\end{table*}

\subsection{Comparison with SOTA}\label{sec:sota}
We compare our best method ``SourceTargetFT+SPL" with state-of-the-art methods in Tables \ref{table:sota_o31} and \ref{table:sota_o65}. For a fair comparison, the results of different methods are grouped according to the models (or backbones) and we compare results within and across groups.

For the Office31 dataset, when the ResNet50 model is used as the backbone, our best method achieves an average accuracy of 90.1\% over six tasks and outperforms state-of-the-art methods. When the transformer based DeiT-small and DeiT-base models are employed, our best method achieves an average accuracy of 91.3\% and 93.1\%, respectively. The performance is higher than those of state-of-the-art methods including the competitive CDTrans \cite{xu2021cdtrans} which employs a triple-branch transformer framework.

On the Office-Home dataset, our best method also achieves the best performance regardless of the employed backbones. In particular, when the DeiT-small model is used, our method achieves an average accuracy of 77.0\% which is significantly higher than the performance of the more complicated end-to-end deep learning based approach CDTrans (74.7\%).
\begin{table}[!t]
	\centering
	{
		\centering
		\caption[]{Classification accuracy (\%) comparison with state-of-the-art methods on the Office31 dataset. * indicates the results are produced by \cite{xu2021cdtrans}.
		}
		\label{table:sota_o31}
		\resizebox{.8\columnwidth}{!}{%
			\begin{tabular}{clccccccc}
				\hline
				Model (Backbone) & Method & \scriptsize{A$\to$W} & \scriptsize{D$\to$W} & \scriptsize{W$\to$D} & \scriptsize{A$\to$D} & \scriptsize{D$\to$A} & \scriptsize{W$\to$A} & Avg \\ \hline
				\multirow{12}{*}{ResNet50} & RTN\cite{long2016unsupervised} & 84.5 & 96.8 & 99.4 & 77.5 & 66.2 & 64.8 & 81.6\\
				& MADA\cite{pei2018multi} & 90.0 & 97.4 & 99.6 & 87.8 & 70.3 & 66.4 & 85.2 \\
				& GTA \cite{sankaranarayanan2017generate} & 89.5& 97.9 & {99.8}& 87.7 & 72.8 & 71.4& 86.5\\
				& iCAN\cite{zhang2018collaborative} & 92.5 & {98.8} & {100.0} & 90.1 & 72.1 & 69.9 & 87.2 \\
				& CDAN-E\cite{long2018conditional} & {94.1} & 98.6 & {100.0} & 92.9 & 71.0 & 69.3 & 87.7\\
				& JDDA\cite{chen2018joint} & 82.6 & 95.2 & 99.7 & 79.8 & 57.4 & 66.7 & 80.2\\
				& SymNets\cite{zhang2019domain} & 90.8 & {98.8} & {100.0} & {93.9} & 74.6 & 72.5 & {88.4}\\
				& TADA \cite{wang2019transferable} &  {94.3} & 98.7 & {99.8} & 91.6 & 72.9 & 73.0 & {88.4}\\
				& MEDA\cite{wang2018visual} & 86.2 & 97.2 & 99.4 & 85.3 & 72.4 & 74.0 & 85.7\\
				& CAPLS \cite{wang2019unifying} & 90.6 & 98.6 & 99.6 & 88.6 & {75.4} & {76.3} & {88.2}\\
				& SPL \cite{wang2020unsupervised} & 92.7 & {98.7} & {99.8} & {93.0} & {76.4} & {76.8} & {89.6} \\
				& SourceTargetFT+SPL & 92.3 & 98.7 & 99.8 & 93.8 & 78.5 & 77.4 & 90.1 \\
				\hline
				ResNet101 & SourceTargetFT+SPL & 93.1 & 99.0 & 99.8 & 94.8 & 79.5 & 78.4 & 90.8\\
				\hline
				
				\multirow{2}{*}{DeiT-small} & CDTrans \cite{xu2021cdtrans} & 93.5 & 98.2 & 99.6 & 94.6 & 78.4 & 78.0 & 90.4 \\
				& SourceTargetFT+SPL & 95.6 & 98.1 & 100.0 & 96.6 & 78.0 & 79.5 & 91.3\\
				\hline
				\multirow{4}{*}{DeiT-base} & SHOT*\cite{liang2020we} & 94.3 & 99.0 & 100.0 & 95.3 & 79.4 & 80.2 & 91.4 \\
				& CGDM* \cite{du2021cross} & 95.3 & 97.6 & 99.8 & 94.6 & 78.8 & 81.2 & 91.2\\
				& CDTrans \cite{xu2021cdtrans} & 96.7 & 99.0 & 100.0 & 97.0 & 81.1 & 81.9 & 92.6 \\
				& SourceTargetFT+SPL & \bf 97.6 & \bf 99.1 & \bf 100.0 & \bf 98.0 & \bf 81.3 & \bf 82.4 & \bf 93.1\\
				\hline
			\end{tabular}%
		}
	}
\end{table}

\begin{table*}[!htbp]
	\centering
	{
		\centering
		\caption[]{Classification accuracy (\%) comparison with state-of-the-art methods on the Office-Home dataset. * indicates the results are produced by \cite{xu2021cdtrans}.}
		\label{table:sota_o65}
		\resizebox{\columnwidth}{!}{%
			\begin{tabular}{clccccccccccccc}
				\hline
				Model (Backbone) & Method & A$\to$C & A$\to$P & A$\to$R & C$\to$A&C$\to$P & C$\to$R&P$\to$A & P$\to$C & P$\to$R & R$\to$A & R$\to$C & R$\to$P & Average \\ \hline
				\multirow{7}{*}{ResNet50} & JAN\cite{long2017deep} & 45.9 & 61.2 & 68.9 & 50.4 & 59.7 & 61.0 & 45.8 & 43.4 & 70.3 & 63.9 & 52.4 & 76.8 & 58.3\\
				& CDAN-E \cite{long2018conditional} & 50.7 & 70.6 & 76.0 & 57.6 & 70.0 & 70.0 & 57.4 & 50.9 & 77.3 & 70.9 & 56.7 & 81.6 & 65.8\\
				& SymNets \cite{zhang2019domain} & 47.7 & 72.9 & 78.5 & 64.2 & 71.3 & 74.2 & 64.2 & 48.8 & 79.5 & {74.5} & 52.6 & 82.7 & 67.6 \\ 
				& TADA \cite{wang2019transferable} & 53.1 & 72.3 & 77.2 & 59.1 & 71.2 & 72.1 & 59.7 & {53.1} & 78.4 & {72.4} & {60.0} & 82.9 & 67.6 \\
				& MEDA\cite{wang2018visual} & {54.6} & 75.2 & 77.0 & 56.5 & 72.8 & 72.3 & 59.0 & 51.9 & 78.2 & 67.7 & {57.2} & 81.8 & 67.0\\
				& CAPLS \cite{wang2019unifying} & {56.2} & {78.3} & {80.2} & {66.0} & {75.4} & {78.4} & {66.4} & {53.2} & {81.1} & {71.6} & 56.1 & {84.3} & {70.6}\\
				& SPL \cite{wang2020unsupervised} & 54.5 & {77.8} & {81.9} & {65.1} & {78.0} & {81.1} & {66.0} & {53.1} & {82.8} & 69.9 & 55.3 & {86.0} & {71.0} \\
				& SourceTargetFT+SPL & 55.9 & 79.3 & 82.3 & 67.1 & 79.6 & 79.4 & 67.0 & 54.8 & 82.1 & 71.6 & 56.7 & 85.7 & 71.8 \\ 
				\hline
				ResNet101 & SourceTargetFT+SPL & 59.2 & 82.5 & 83.8 & 71.3 & 80.4 & 81.8 & 68.1 & 55.4 & 83.7 & 73.4 & 58.3 & 87.7 & 73.8\\
				\hline
				\multirow{2}{*}{DeiT-small} & CDTrans \cite{xu2021cdtrans} & 60.6 & 79.5 & 82.4 & 75.6 & 81.0 & 82.3 & 72.5 & 56.7 & 84.4 & 77.0 & 59.1 & 85.5 & 74.7\\
				& SourceTargetFT+SPL & 63.8 & 82.9 & 85.2 & 76.2 & 84.4 & 83.8 & 75.2 & 61.5 & 85.8 & 77.4 & 60.0 & 87.2 & 77.0 \\
				\hline
				\multirow{4}{*}{DeiT-base} & SHOT*\cite{liang2020we} & 67.1 & 83.5 & 85.5 & 76.6 & 83.4 & 83.7 & 76.3 & \bf 65.3 & 85.3 & 80.4 & 66.7 & 83.4 & 78.1 \\
				& CGDM* \cite{du2021cross} & 67.1 & 83.9 & 85.4 & 77.2 & 83.3 & 83.7 & 74.6 & 64.7 & 85.6 & 79.3 & \bf 69.5 & 87.7 & 78.5 \\
				& CDTrans \cite{xu2021cdtrans} & 68.8 & 85.0 & 86.9 & \bf 81.5 & \bf 87.1 & \bf 87.3 & \bf 79.6 & 63.3 & 88.2 & \bf 82.0 & 66.0 & \bf 90.6 & \bf 80.5 \\
				& SourceTargetFT+SPL & \bf 69.4 & \bf 85.1 & \bf 87.9 & 81.2 & 85.7 & 86.5 & 78.0 & 64.4 & \bf 89.0 & 81.0 & 66.7 & \bf 90.6 & \bf 80.5 \\
				\hline
			\end{tabular}%
		} 
	}
\end{table*}

\section{Conclusion}
In this paper, we propose an effective fine-tuning mechanism for improved deep features for UDA. From the experimental results of our empirical studies, several conclusions can be drawn: properly fine-tuned deep features can improve the performance of feature transformation based domain adaptation approaches such as SPL \cite{wang2020unsupervised}; it is usually beneficial to fine-tuning the models with both labelled source data and pseudo labelled target data; special care needs to be taken for batch normalisation layers in the deep CNN models during fine-tuning; and the performance gap between transformer and CNN based approaches is attributed to the improved deep features extracted from transformer based models.

\bibliography{egbib}
\end{document}